\documentclass[10pt,twocolumn,letterpaper]{article}

\usepackage{cvpr}              

\usepackage{graphicx}
\usepackage{amsmath}
\usepackage{amssymb}
\usepackage{booktabs}

\usepackage[accsupp]{axessibility}

\usepackage[pagebackref,breaklinks,colorlinks]{hyperref}

\usepackage[capitalize]{cleveref}
\crefname{section}{Sec.}{Secs.}
\Crefname{section}{Section}{Sections}
\Crefname{table}{Table}{Tables}
\crefname{table}{Tab.}{Tabs.}

\def\cvprPaperID{13} 
\def\confName{AgriVision}
\def\confYear{2023}

\begin{document}

\title{Mushroom Segmentation and 3D Pose Estimation from Point Clouds \\ using Fully Convolutional Geometric Features and Implicit Pose Encoding}

\author{George Retsinas, Niki Efthymiou, Petros Maragos\\
\\
School of Electrical and Computer Engineering, National Technical University of Athens, Greece\\
{\tt\small \{gretsinas,nefthymiou\}@central.ntua.gr,\, maragos@cs.ntua.gr
}
}
\maketitle

\begin{abstract}
Modern agricultural applications rely more and more on deep learning solutions. 
However, training well-performing deep networks requires a large amount of annotated data that may not be available and in the case of 3D annotation may not even be feasible for human annotators. 
In this work, we develop a deep learning approach to segment mushrooms and estimate their pose on 3D data, in the form of point clouds acquired by depth sensors. 
To circumvent the annotation problem, we create a synthetic dataset of mushroom scenes, where we are fully aware of 3D information, such as the pose of each mushroom. 
The proposed network has a fully convolutional backbone, that parses sparse 3D data, and predicts pose information that implicitly defines both instance segmentation and pose estimation task. 
We have validated the effectiveness of the proposed implicit-based approach for a synthetic test set, as well as provided qualitative results for a small set of real acquired point clouds with depth sensors.  
Code is publicly available at \url{https://github.com/georgeretsi/mushroom-pose}.

\end{abstract}

\section{Introduction}
\label{sec:intro}

The need for autonomous harvesting is increasing in agriculture due to labor scarcity and the growing population. Automated harvesting robots require accurate 3D representation and pose estimation to navigate through crops, locate fruits, and harvest them without damaging the plants or the fruits. Due to challenging lighting conditions, occlusions, and plant growth, there is a growing need to develop 3D vision approaches to provide accurate segmentation and pose estimation. These technologies could help farmers to increase production and harvesting effectiveness and enhance the overall quality of their products.
The case of interest in this work is the commercial harvesting of the white cultivated mushroom, Agaricus bisporus, on industrial mushroom farms. 

Starting from Ciarfuglia et al. \cite{ciarfuglia2022pseudo} facing the limited available annotated data for fruits in orchards, they addressed the challenge of generating pseudo-labels for agricultural robotics applications. They proposed a method for generating high-quality pseudo-labels using a combination of self-supervised learning and transfer learning.
Le Louedec et al. in \cite{le2020segmentation} presented a CNN-based method for segmenting and detecting broccoli heads from 3D point clouds. They assessed their approach to broccoli fields by evaluating both semantic and instance segmentation and providing a qualitative analysis. 
In \cite{wang20223d}, Wang et al. focused on instance segmentation in 3D point clouds. They trained PartNet giving as input the point cloud of lettuce and taking the segmentation of each leaf as output, using point clouds from real and synthetic datasets. 

As far as we know, there has been little research on mushrooms' 3D posture assessment. Qian et al.~\cite{qian2020real} presented an object recognition and localization method for an oyster mushroom harvesting robot that blends detection data from a neural network with depth data from an RGB-D camera. The SSD object detection algorithm and depth pictures based on binocular and structured light principles were utilized in this technique to determine the precise position of the identified item in the 3D environment, ensuring real-time performance. In \cite{baisa2022mushrooms}, the authors proposed a method for recognizing and segmenting mushrooms in 3D space using RGB and depth information. The mushrooms are recognized using a mix of active contouring and the circular Hough transform, and their 3D location and orientation are approximated using registration methods using a template mushroom model. 

In this work, we developed a 3D deep network for separating mushrooms and estimating their pose from point cloud inputs. 
Specifically, we assume that the input of our system is a mushroom scene, in the form of a point cloud, obtained from depth sensors such as RealSense active-stereo cameras, possibly in a multi-view setting to enhance point cloud reconstruction quality.
To train our system, a pipeline for creating synthetic mushroom scenes was developed. 
This way, we acquire all the annotations required to train a well-performing system on the considered  tasks.
Architecture-wise, we relied on Fully Convolutional Geometric Features (FCGF)~\cite{choy2019fully} and we used the same 3D convolutional network that processes point clouds as our backbone.
On top of this backbone, we added a small fully-connected network to predict task-relevant variables at each point of the point cloud.
Emphasis was given on the per-point predicted variables to achieve the best possible performance, leading to the definition of the proposed implicit pose encoding that enables us to achieve notably high performance. 

The contributions of this work are summarized below:\\
    $\bullet$
     We developed a synthetic dataset of point clouds that describe mushroom scenes to address the lack of annotated data. These point cloud scenes have several augmentation steps to simulate realistic point clouds gathered from depth sensors. The generated synthetic data can help us train deep 3D networks for this task. \\
    $\bullet$
    We proposed a point-level implicit pose encoding that describes pose information in an indirect manner. 
    An indicative variable of this encoding, used for estimating rotation, is a singular value for each point that denotes how close we are to the top of the cap. 
    This encoding is correlated with both tasks at hand and considerably assists the convergence of the network.  
    \\
    $\bullet$
    Instance segmentation is not explicitly performed through the network, but a clustering step is used to separate mushrooms to avoid complex architectures with multi-prediction options.\\
    $\bullet$
    We validate the effectiveness of our method in a synthetic test and provide qualitative results over real data, aiming to highlight the synthetic-to-real adaptation.   

The remainder of this paper is structured as follows: 
Section 2 presents the synthetic point cloud generation step, while Section 3 describes the proposed pipeline for mushroom detection and 3D pose estimation. Finally, Section 4 reports our experimental evaluation both on a synthetic dataset and real data and Section 5 concludes our work.



\section{Creating Synthetic Point Clouds}
\label{sec:synthetic}

\begin{figure}[h]
    \centering
    \includegraphics[width=.40\linewidth]{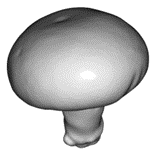} 
    \caption{3D mesh of the mushroom template.}
    \label{fig:template}
\end{figure}

The goal of this work is to modify a deep network in order to perform operations in point clouds of mushroom scenes. 
Nonetheless, even fine-tuning an existing pipeline requires annotated data. 
Such information is not available for our task.
Furthermore, annotating point clouds and especially providing 3D pose annotation is impractical for large scale data collections.
To this end, we focused our efforts on creating a pipeline for generating diverse synthetic point clouds.
The building block of our mushroom scene creation pipeline is a 3D mushroom mesh, as shown in Figure~\ref{fig:template}.
The creation of mushroom scenes can be summarized as the random placement of transformed versions of this template over a ground plane. 

Since we want non-trivial variations over the mushrooms and the generated scene, we followed a series of augmentations, as described in what follows.

\begin{figure*}[t]
    \centering
    \resizebox{\linewidth}{!}{
    \begin{tabular}{ccc}
     \includegraphics[width=.31\linewidth]{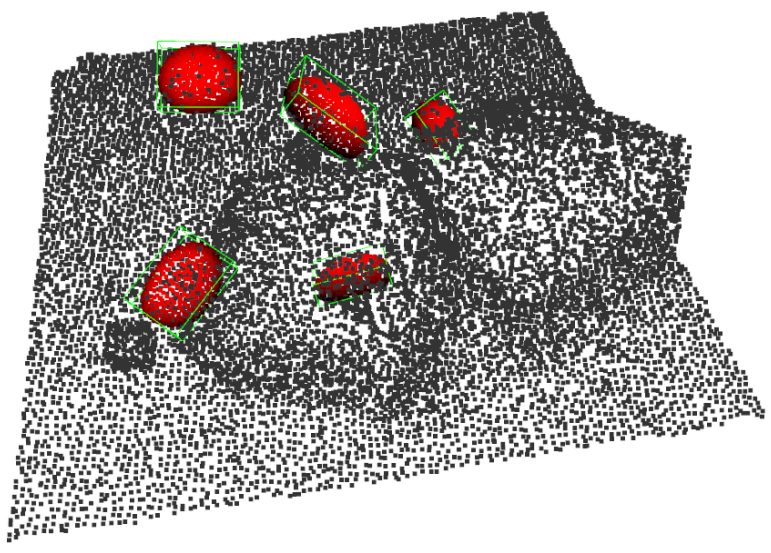} &
     \includegraphics[width=.31\linewidth]{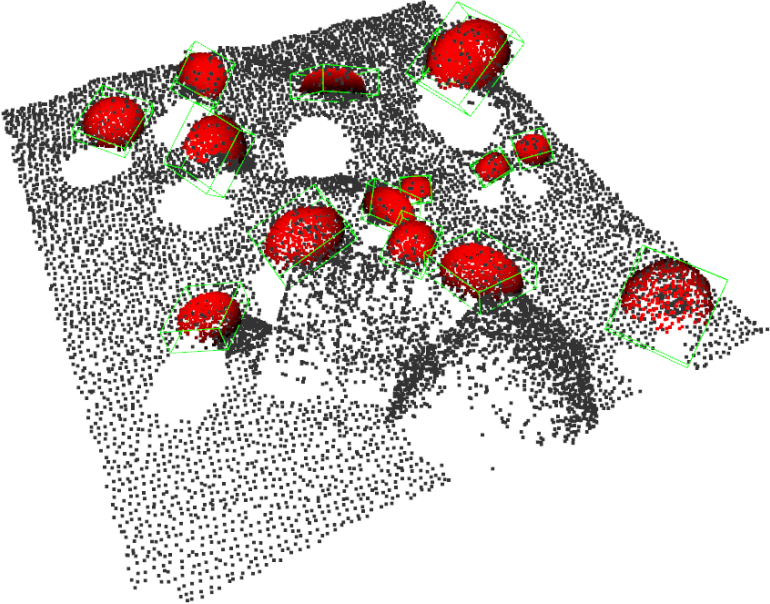} &
     \includegraphics[width=.31\linewidth]{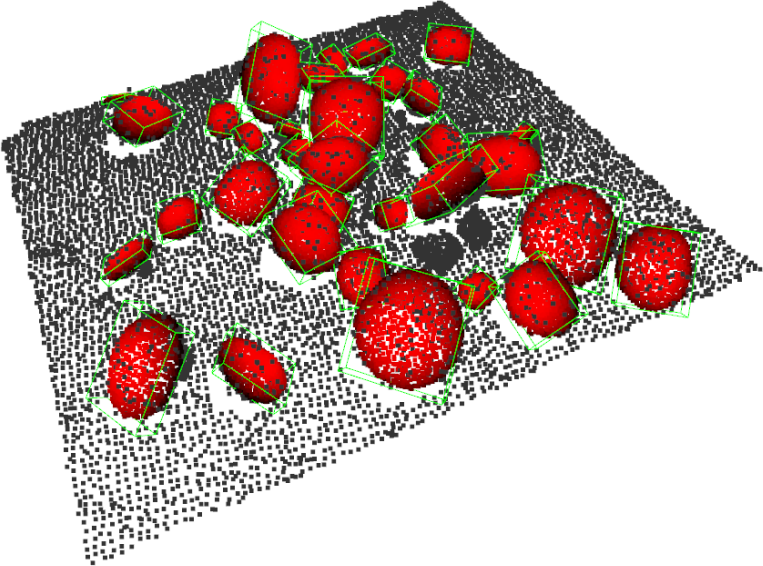} \\
     (a) 5 mushrooms & (b) 15 mushrooms & (c) 35 mushrooms 
    \end{tabular}
    }
    \caption{Generated scenes of synthetic point clouds. Distractors are clearly visible in the first two images. Mushrooms regions are highlighted with red color.}
    \label{fig:synthetic}
\end{figure*}

\textbf{Ground Augmentations:} We use realistic non-smooth ground planes. \\
    $\bullet$
    Two different soil meshes were considered with unique terrain deformations.  \\
    $\bullet$
    We further deform these ground meshes. A local deformation approach was used where a subset of points was randomly selected, and a random translation magnitude was assigned to them. 
    We translated these points along their normals. 
    Neighboring points were also translated using an interpolation step.\\
    $\bullet$
    Random ``distractors'' were added to the scenes. These distractors are either cubes or cones with random position, scale and rotation. 
    Their goal is to introduce ``foreign'' elements in the scenes and increase robustness of the mushroom detection.

\textbf{Mushroom Augmentations:}
We randomly select a number $K$ of mushrooms to be placed on the ground mesh. The mushroom template of Figure~\ref{fig:template} is used in this step. The augmentation sub-steps for each mushroom are summarized below. \\
    $\bullet$
    Scale the mushroom template within the range $s \in [0.5, 1.5]$. A finer per-axis re-scaling step is then used for extra mushroom variability by randomly selecting a factor in the range $a \in [0.8, 1.2]$ (e.g., $s_z = a_x s$). \\
    $\bullet$
    Rotate the template using random axis angles over a constrained set.
    Specifically, rotation over the x- and y-axis was constrained to the range $[-45^\circ, 45^\circ]$. Due to symmetry over the z-axis, we choose to leave this rotation unconstrained.\\
    $\bullet$
    Apply local deformations in the mushroom mesh (along the surface normals) without significantly altering its surface. For this step, we use the same rationale for the local ground deformation.\\
    $\bullet$
    Translate the mushroom template anywhere on the ground mesh (considering only translation over the xy-axes). The mushroom is also translated along the z-axis in order to have its bottom point in the proximity of the ground plane.\\
    $\bullet$
    Basic collision checking is applied. If a newly created mushroom collides with an existing one, we discard the new mushroom and create a new one, iteratively, until no collision is detected.

\textbf{Scene Augmentations:} Generic scene augmentation steps that apply on both ground, distractor and mushroom objects.\\
    $\bullet$ To create the final point cloud, use a slightly different sampling number of points and different voxel sizes ($\times0.8$ - $\times1.2$) for the subsequent down-sampling in order to provide extra variability in the created scenes.\\
    $\bullet$
    Lastly, to simulate realistic collected point clouds, a hidden point removal step is performed~\cite{katz2007direct}. This step approximates the visibility of a point cloud from a given view and removes occluded points. Possible views are randomly selected, using a constrained radius range and a constrained set of axis angles that do not considerably diverge from an overhead viewpoint.

The ``power'' of these synthetic scenes is the ability to obtain 3D annotations automatically. 
We considered the following annotations:\\
    $\bullet$ A binary label is used to annotate if a point of the scene belongs to a mushroom template or not. Specifically, we only label mushroom points belonging to the mushroom cap since the stem is often hidden and irrelevant to a mushroom picking application. \\
    $\bullet$ The center of each mushroom as a 3D point.\\
    $\bullet$ Oriented 3D bounding box of each mushroom. \\
In the forthcoming sections, we will discuss extra possible annotations, as by-product of the aforementioned ones, that act as auxiliary pose encodings.

Examples of the generated scenes with increasing number of mushrooms are presented in Figure~\ref{fig:synthetic}. 

\section{Proposed Approach}
\label{sec:proposed}

\begin{figure*}[t]
    \centering
     \includegraphics[width=.9\linewidth]{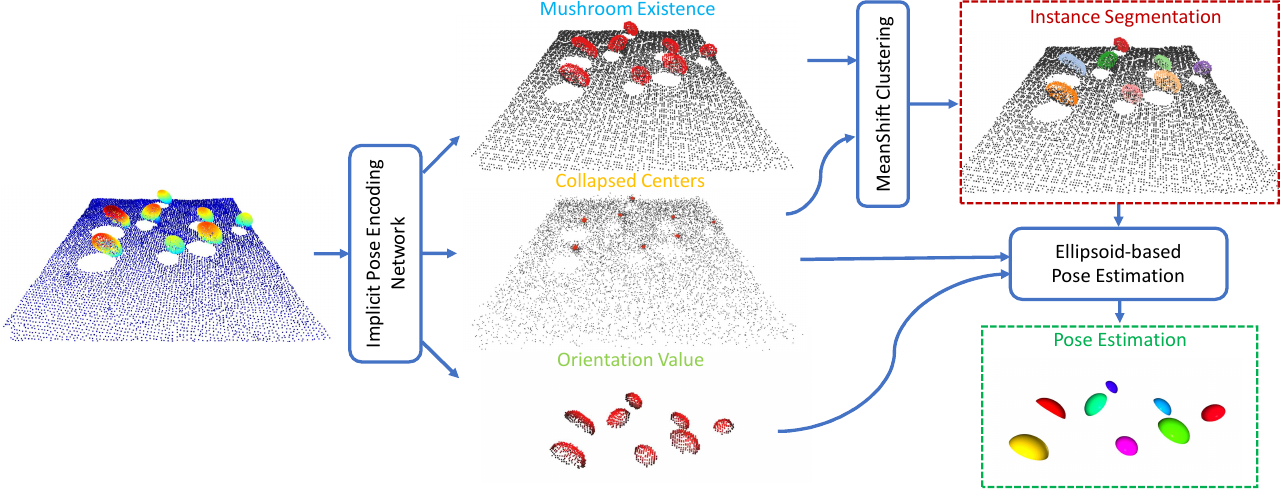} 
    \caption{Overview of the proposed system. Given a point cloud input of a mushroom scene, the proposed deep network predicts the three categories of task-relevant information. Using a mode-seeking clustering over the predicted centers we can provide the instance segmentation result. Then each mushroom region is processed as an ellipsoid structure and the corresponding 3D pose is estimated.}
    \label{fig:overview}
\end{figure*}

In this work, we develop a method that performs both instance segmentation and pose estimation tasks by relying on a set of per-point features, referred  to as implicit  pose encoding. These task-related variables are predicted through a 3D convolutional network, inspired  by Fully Convolutional Geometric Features~\cite{choy2019fully}.
The overview of our method can be found in Figure~\ref{fig:overview}. The network architecture, the pose encoding and the underlying functionalities in order to perform segmentation and pose estimation will be described in detail in the following sections.

\subsection{Fully Convolutional Geometric Features}

The backbone of the proposed system is a sparse 3D convolutional network, proposed by Choy et al.~\cite{choy2019fully}. 
This network was developed to extract geometric features, referred to as Fully Convolutional Geometric Features (FCGF), from 3D point clouds.
The authors validated the effectiveness of their method in registration tasks (for both indoor and outdoor data), where a point matching step using the 3D features was required. 

The core operation of this network is 3D convolution over sparse data. 
The point cloud input is organized as a sparse tensor after a voxel down-sampling step. 
This voxel-based down-sampling operation essentially creates a 3D grid, which is expected to be considerably sparse, as in any quantization of a 3D point structure.    
Features are then extracted by a single forward pass of this fully-convolutional 3D network. 
The sparse representation of the data assists us in efficiently performing these demanding 3D convolutions, using an appropriate library that supports sparse computations (Minkowski Engine~\cite{choy20194d}).

The authors of~\cite{choy2019fully} highlighted the effectiveness of metric learning losses with a ``hard" example sampling process over contrastive or triplet loss setups.
Due to the nature of our task, we did not explore such metric losses, and we used the 3D network of FCGF as a feature extractor backbone, initializing our own model with the provided pre-trained model by the authors of ~\cite{choy2019fully}.

In this work, the proposed architecture model uses a simple, fully connected network of three linear layers, intervened by ReLU non-linearities, on top of the FCGF backbone that transforms the per-point 3D features into task-relevant predictions. 
We will discuss the possible predictions, referred to as implicit pose encoding, in the following section.

\subsection{Proposed Implicit Pose Encoding}

In this section, we will describe the possible task-oriented predictions of our 3D model with respect to our multifacet task: segmentation and 3D pose estimation.

As we will show in the experimental section, we have discovered that trying to learn a very specific regression target, such as the center of the mushroom, at point-level may lead to subpar performance. 
Specifically, the 3D network should assign the same regression value (e.g., pose parameters) to a subset of the points (i.e., the points belonging to a specific mushroom). 
Such a straightforward formulation of our problem proved difficult to train in practice.
To this end, we devised a set of auxiliary variables that can implicitly define the requested tasks, dubbed as implicit pose encoding.

In more detail, we trained our network to simultaneously predict point-level information relevant to both segmentation and pose estimation tasks. 
Specifically, for each point of the point cloud, we estimate the following set of variables:\\
    $\bullet$
    \emph{existence value:} we simply predict whether a point belongs to a mushroom or the background. We consider only points belonging to mushroom caps as foreground.\\
    $\bullet$
    \emph{residual center:} for each mushroom point $p$, we predict its corresponding mushroom center $c$ in a residual rationale, i.e., the target 3D vector is defined as the difference $\mathbf{c} - \mathbf{p}$. This residual approach helped convergence, giving a more intuitive, point-based formulation compared to a fixed regression target. \\
    $\bullet$
    \emph{orientation value:} a single value metric denoting how close a point is to the top of the cap that implicitly defines a rotation vector. Each point is assigned one value from zero to one, where zero corresponds to the base of the cap (the wider part) and one corresponds to the top of the cap.

We should highlight that these variables implicitly define the requested information. 
However, they do this in a self-referenced way for each point, essentially addressing the following question: ``how can we get a glimpse of requested pose information when we are at this specific point?''. Such an approach proved very effective in practice and notably assisted the training procedure in converging to well-performing solutions.

Training is performed using a multi-task loss. In this context, each ``task'' corresponds to one of the three aforementioned categories of variables. Mushroom existence values is trained through a binary cross entropy loss.
Center and orientation information are trained with mean squared error loss.
Since residual center targets typically take small values, an increased weight is considered for this subtask ($\times 100$).
Center residuals and pose values are zero-ed for non-mushroom points.
The overall loss is the summation of the aforementioned individual losses.

In what follows, we describe how we utilize these variables to first perform instance segmentation of mushrooms and then pose estimation of each individual mushroom. 

\begin{figure}[t]
    \centering
    \begin{tabular}{c}
     \includegraphics[width=.75\linewidth]{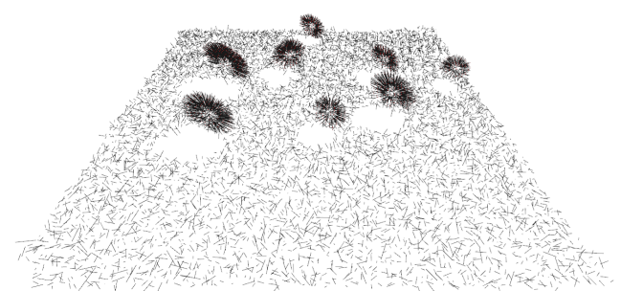} \\
     \includegraphics[width=.75\linewidth]{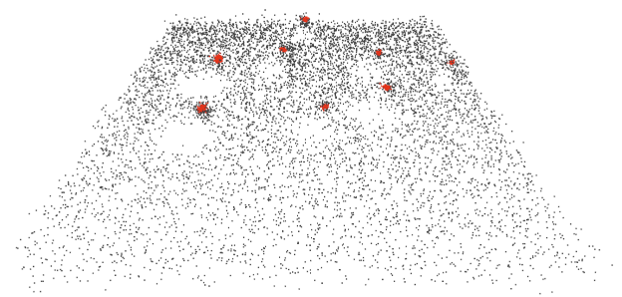}
    \end{tabular}
    \caption{The residual center information represented as translation vector (top) and the corresponding transformation that creates dense regions around mushroom centers (bottom). Note how the mushroom points in the top image  are directed towards the center, while background points have smaller random displacements.}
    \label{fig:collapse}
\end{figure}

\subsection{Instance Segmentation}
\label{sec:instance}

We have defined how to predict several useful variables, but we have not described how they will be used in the context of our work. 
The mushroom existence prediction can provide a faithful foreground/background segmentation. 
Nevertheless, instance segmentation requires that each individual mushroom is assigned to a different label.
Thus, we have developed two clustering-based alternatives: \\
$\textbf{1)}$ We assume that mushrooms do not touch each other. Therefore, between each mushroom, there is a gap with no annotated mushroom points. 
    Following this assumption, mushrooms can be separated with a density clustering algorithm, such as DBSCAN (~\cite{ester1996density}). 
    Nonetheless, this assumption does not hold for every possible mushroom scene. 
    In fact, it is common for neighboring mushrooms, growing in mushroom farms, to touch each other. 
    To overcome this issue, we enforce the aforementioned assumption by assigning a non-mushroom label to mushroom points that are close together in the target labels of the synthetic scenes. 
    Specifically, for each point we find the two closest mushroom centers and we assign as non-mushrooms points that have similar distances from these centers (i.e., we discard points using the condition $\| 1 - d_1/d_2 \| < .25$ where $d_1$, $d_2$ are the distances to the two closest mushroom centers).
    This approach relies only on the mushroom existence value, given the proposed labeling modification that spatially separates mushrooms regions.\\
$\textbf{2)}$ We utilize the center information, along with the mushroom existence values. Given a well-performing network, we expect that each mushroom point will ``collapse'' to its center. 
    Such behavior will create very dense regions of points that correspond to individual mushrooms, as seen in Figure~\ref{fig:collapse}. 
    According to this rationale, we selected a mode-seeking clustering algorithm, such as the MeanShift algorithm~\cite{cheng1995mean}, for clustering points into separate mushrooms. 

\subsection{3D Pose Estimation}

Given a well-performing instance segmentation step, one could perform pose estimation using a template registration scheme (e.g., Iterative Closest Point algorithm) over individual mushroom regions. 
However, such iterative template matching/registration approaches introduce non trivial overhead when performed for each mushroom separately. 
To avoid such computational overhead, we make the assumption that an ellipsoid structure can approximate the mushroom cap. 

Following an ellipsoid formulation, we can derive the rotation parameters using only the per-point singular orientation values.
Formally, the ``rectified" points should, after applying the rotation matrix, will correspond to the network's orientation predictions, which in practice is the normalized projections in the z-axis, or formally:
\begin{equation}
[R (p - c)^T]_z = \vec{e}_z R (p - c)^T = y_o  s_z
\end{equation}
, where $y_o$ is the orientation predictions provided by the network, $s_z$ the scale over z-axis and $\vec{e}_z$ the unitary row vector with respect to z-axis ($[0\, 0\, 1]$).

Assuming the the rotation over $z$-axis is irrelevant due to symmetry, we consider the rotation matrix as the product of the basic rotations over x- and y-axes, i.e., $R = R_y(\theta_y) R_x(\theta_x)$.

Therefore, for the set of points $P$, belonging to a potential mushroom cap, and their corresponding orientation predictions $Y_o$ ($Y_o$ is the column vector of all the predicted $y_o$),
we must approximate the following linear system with respect to a 3D vector $[\lambda_1,  \lambda_2, \lambda_3]$ that corresponds to the last row of the rotation matrix $R$ scaled by $s_z$:

\begin{align}
   (P - c)
    \begin{bmatrix} 
     \lambda_1 \\  \lambda_2 \\ \lambda_3
    \end{bmatrix} =
    (P - c)
    \begin{bmatrix} 
    -\sin(\theta_y) / s_z \\ \sin(\theta_x)\cos(\theta_y)  / s_z\\ \cos(\theta_x)\cos(\theta_y)  / s_z
    \end{bmatrix} = Y_o
\end{align}

We can approximate the values $\lambda_i/s_z$ using an ordinary least squares solution and then calculate the relevant ellipsoid parameters:
\begin{align}
s_z = 1 / \sqrt{\lambda_1^2 + \lambda_2^2 + \lambda_3^2} {\label{eq:sz}}\\
\sin(\theta_x) = \lambda_2/ \sqrt{1 - \lambda_1^2}\,\, , \,\,
\sin(\theta_y) = - \lambda_1  s_z
\end{align}
From these equations (and the assumption of an angle range of $[-\pi/2, \pi/2]$ so that cosine values are always positive), we can fully calculate the rotation matrix $R$.

In a nutshell, the $y_o$ predictions are sufficient to estimate how the cap has oriented with respect to its initial upright position. We take into account the symmetry of the cap and the resulting rotation matrix, controlled only by the rotations over x- and y-axes, is analytically estimated.

Finally, we focus on calculating the scale parameters. 
Due to symmetry, we assume $s_x = s_y = s$. To avoid solutions where $s_z$ is much smaller or much larger than $s$, we assume that the ellipsoid has a constrained shape of specific proportions where $s_z = 2 s / 3$. This ratio was selected with respect to the initial template mushroom.
According to this formulation, we discard the previous estimation of $s_z$ and we calculate a ``global'' using the ellipse equations.
Since we want to estimate a singular scale value, a least squares solution is trivially computed. 
Note that Eq.~\ref{eq:sz} is sufficient to compute an estimation of $s$ if we consider a predefined relation between $s_x$, $s_y$ and $s_z$. 
Nevertheless, we have seen that redefine $s$ through the ellipsoid equations works better in practice.

\section{Experimental Evaluation}
\label{sec:experimental}


For our experiments, we trained the proposed system for 20K iterations, with a new random scene generated at each iteration. The number of mushrooms for each scene was randomly selected from the range $[5, 45]$. An Adam optimizer was used along with a multistep scheduler.
For the ablation study, we created a synthetic dataset of 50 mushroom scenes using our synthetic point cloud generation pipeline. The total number of mushrooms in this validation dataset is 529.

Before attempting to measure pose estimation, we should find the correspondences between existing mushrooms and their potential predicted candidates. 
Thus, our initial evaluation is performed through retrieval metrics; Mean Average Precision (MAP) is reported, with an overlap Intersection-over-Union (Iou) threshold dictating successful detections. 
Since the two sub-tasks of detection and pose estimation are intervened, such a retrieval-based take on the problem could provide a quantitative evaluation. 

To perform finer comparisons, we also report a scale error and a 3D orientation error. We only report these metrics for ``good" detection, i.e., detected mushrooms with IoU overlap above a defined threshold.
Specifically, the scale error is defined as relative error: $|s_{predicted} - s_{real}|_1 / s_{predicted}$, while the orientation error is defined as a cosine similarity metric and its corresponding angle error.  

Moreover, we also provide qualitative results of our method in realistic point cloud data, acquired by RealSense active-stereo cameras. Specifically, we considered multi-view settings, with a rotating camera system, to reconstruct mushroom surfaces as faithfully as possible.

\begin{figure*}[h!]
    \centering
    \begin{tabular}{ccc}
    \includegraphics[width=.31\linewidth]{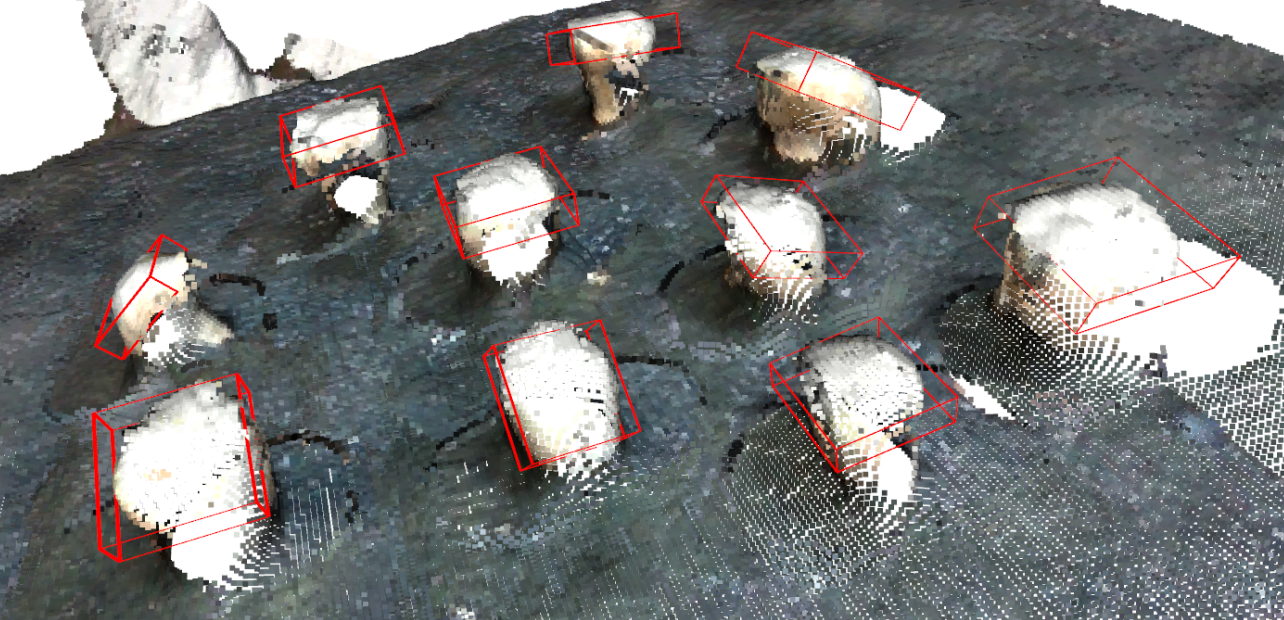} &
     \includegraphics[width=.31\linewidth]{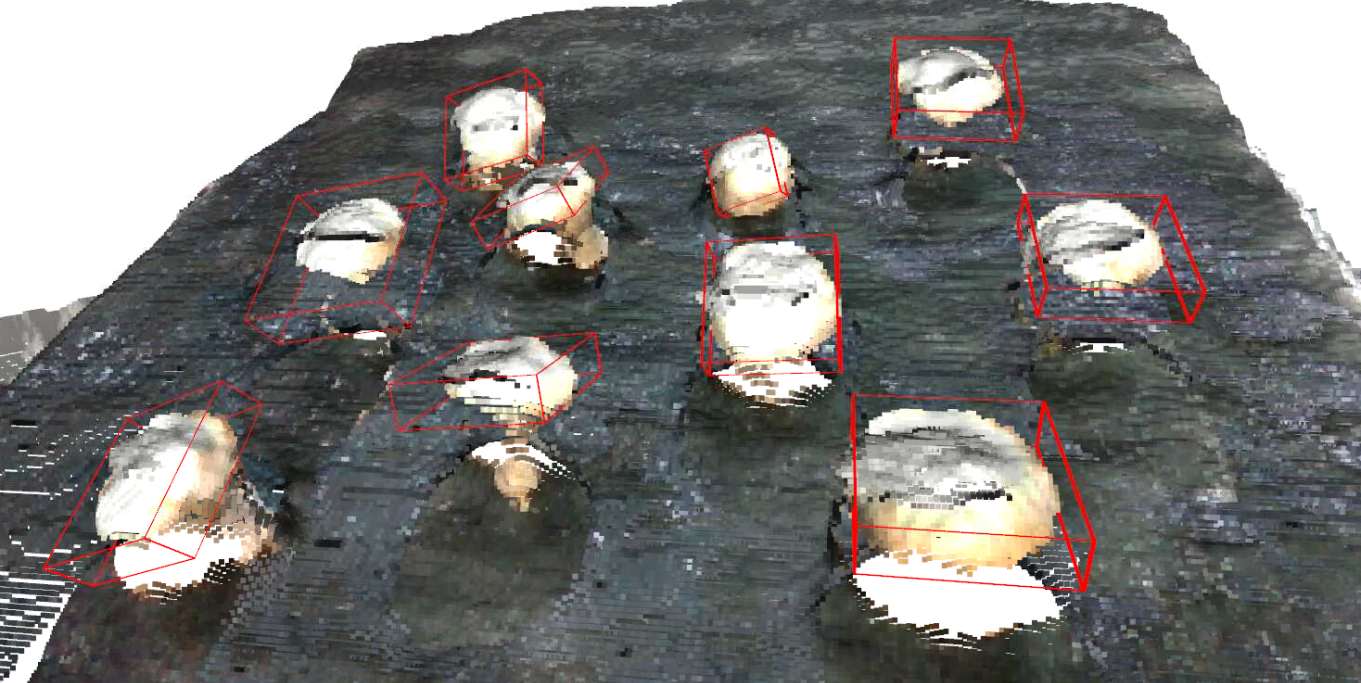} &
     \includegraphics[width=.31\linewidth]{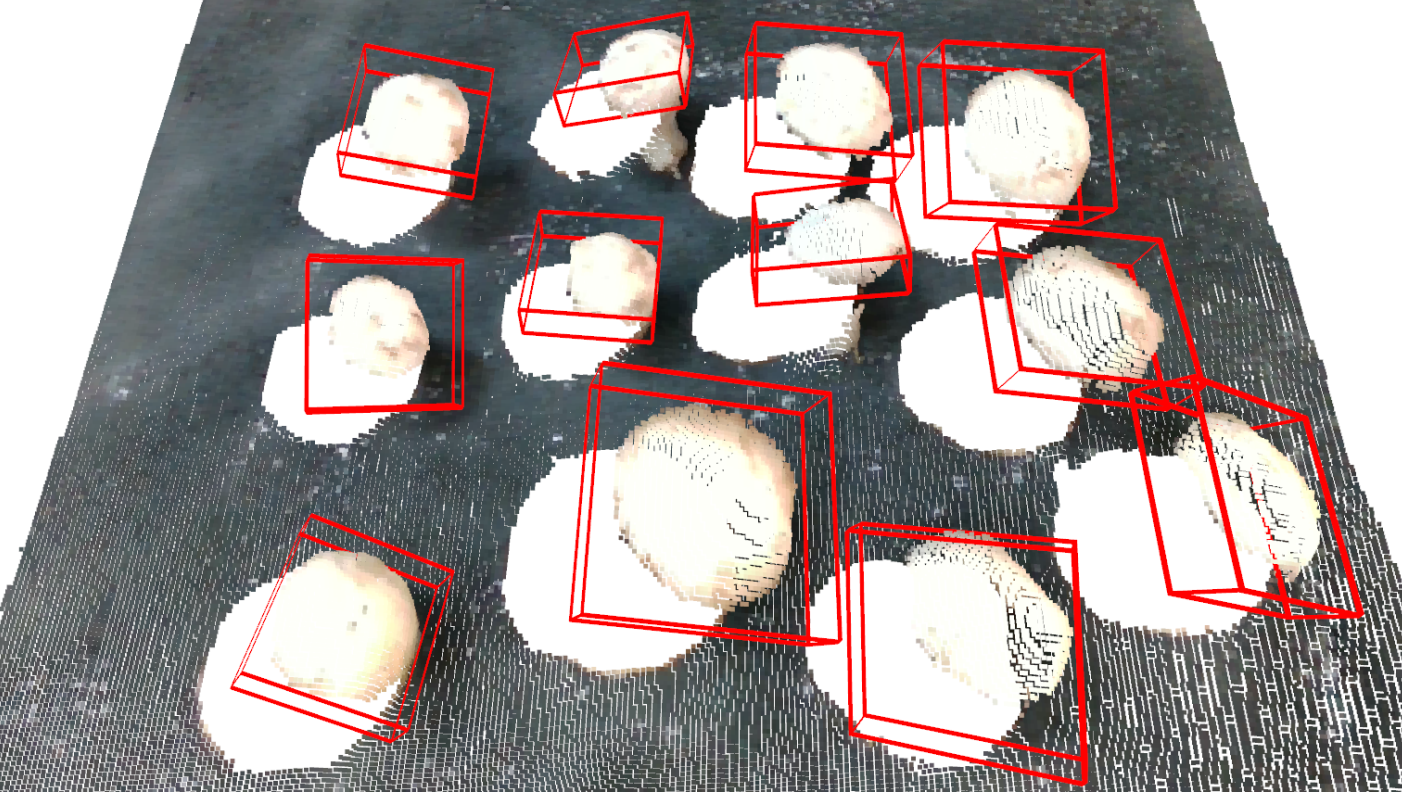} \\
    \includegraphics[width=.31\linewidth]{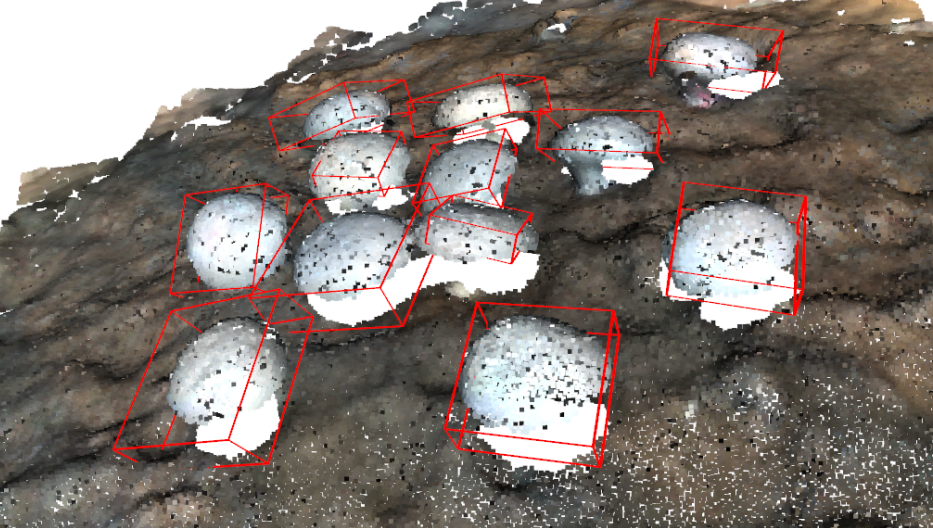} &
     \includegraphics[width=.31\linewidth]{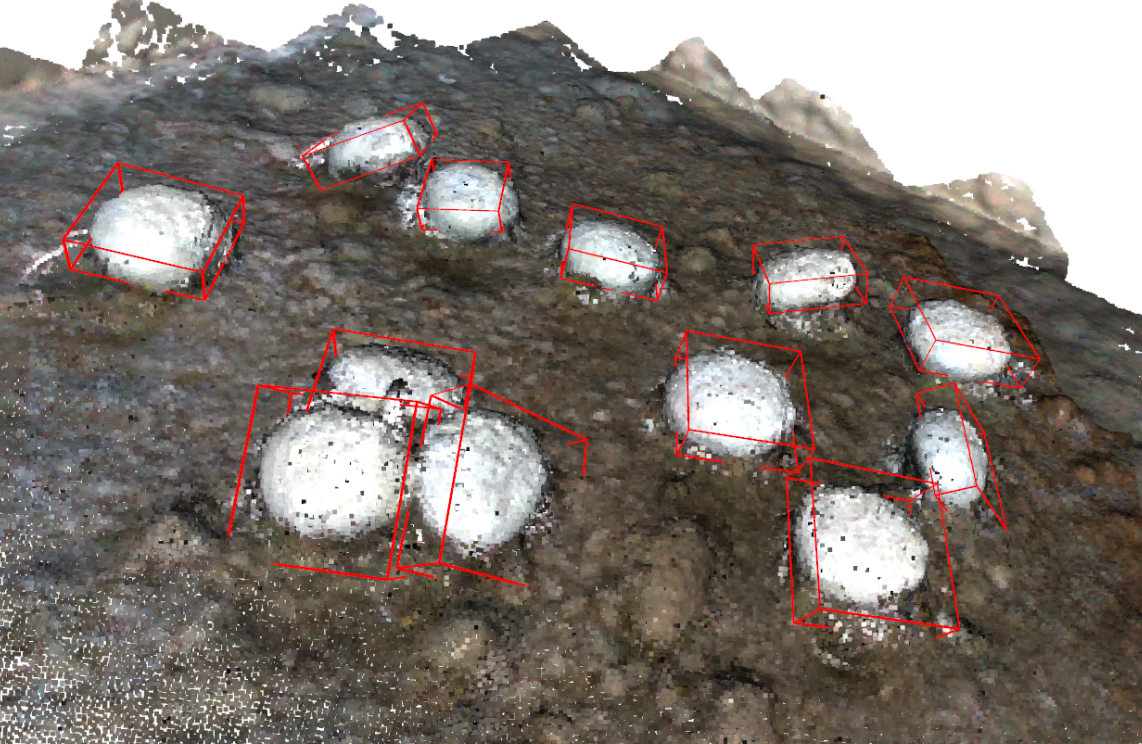} &
     \includegraphics[width=.31\linewidth]{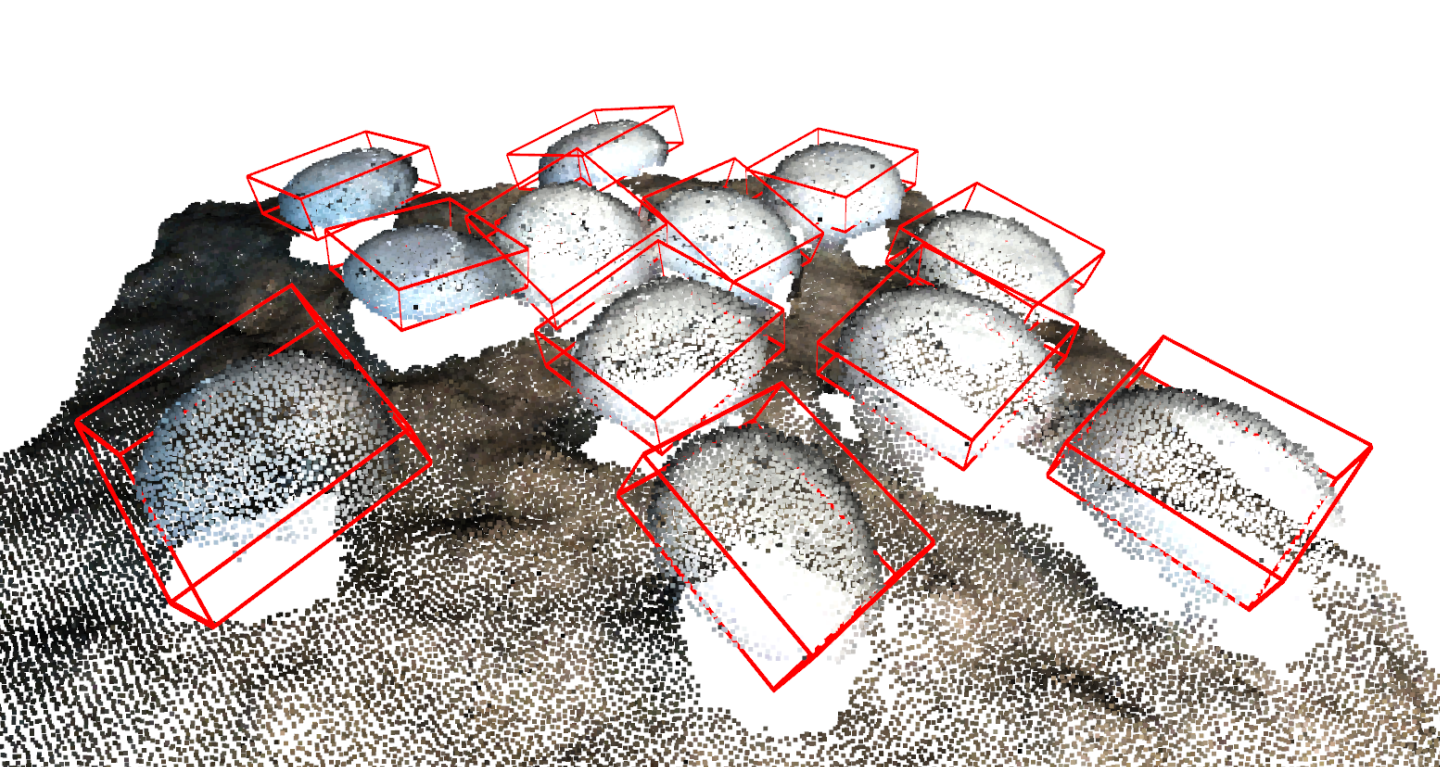}
    \end{tabular}
    \caption{Qualitative examples of segmentation and pose estimation: first row corresponds to data acquired from a setup of two depth RGB-D cameras, while second row correspond to multi-view data of 18 view from a rotating camera system. Top-right image is a single view example. We use a red oriented bounding box to denote the pose of each mushroom. Zoom in for details.}
    \label{fig:vis_bboxes}
\end{figure*}

\subsection{Ablation over Synthetic Data}

We start our exploration by evaluating a basic template matching approach. Specifically, we used the RANSAC algorithm over matches between the 3D features of the scene and features of the template (Fig.~\ref{fig:template}).
The interesting thing here is to validate if the re-trained features and the final predictions of the proposed system can enhance such a typical template matching process.
The results of this experiment are summarized in Table~\ref{tab:ransac}, where the fine-tuned 3D features provide a considerable boost.
However, the final pose-related predictions of our system lead to under-performance. 
This can be attributed to the feature-matching step of the RANSAC system, since the reduced set of 5 dimensions can instigate a plethora of redundant matches that further hinder this stochastic process.

\begin{table}[h]
  \centering
  \resizebox{.85 \linewidth}{!}{
  \begin{tabular}{@{}lc@{}}
    \toprule
    Features & MAP @ 25\% IoU \\
    \midrule
    pre-trained FCGF & 75.74\% \\
    fine-tuned FCGF & 87.40\% \\
    fine-tuned FCGF + pose encoding & 80.17\% \\
    \bottomrule
  \end{tabular}
  }
  \caption{Comparison of RANSAC-based algorithms when using the 3D features, either from the pre-trained network or from our fine-tuned version. Serves as a baseline for forthcoming experiments.}
  \label{tab:ransac}
\end{table}

Next, we explore how the concept of indications assisted the whole system instead of regressing straightforwardly to target values. 
Specifically, we considered a version where the center information is trained explicitly, and thus, each point of the same mushroom should have the same center prediction. 
Also, we considered a version where orientation information is encoded as an orientation vector to be predicted by each point belonging to the mushroom. 
The results of this exploration are presented in Table~\ref{tab:clues}.
We can see that when we did not use our implicit encoding, the system considerably under-performs.
In fact, we reported increased training loss that indicated convergence issues.
However, the proposed implicit definitions provide a close-to-perfect detection score, outperforming also the RANSAC variants of Table~\ref{tab:ransac}.

\begin{table}[h]
  \centering
  \resizebox{.85\linewidth}{!}{
  \begin{tabular}{@{}lc@{}}
    \toprule
    Predictions & MAP @ 25\% IoU \\
    \midrule
    explicit center + implicit orientation & 25.35\% \\
    implicit center +  explicit orientation & 43.90\% \\
    implicit center + implicit orientation & 99.57\% \\
    \bottomrule
  \end{tabular}
  }
  \caption{Impact of using implicit encoding. We considered both center (without residual formulation) and orientation (as an orientation vector) explicit alternatives.}
  \label{tab:clues}
\end{table}

Lastly, we explored the two different instance segmentation approaches, namely the density clustering over the actual points or the clustering over dense ``collapsed'' centers, as described in Section~\ref{sec:instance}.
We denote each approach with the name of its core clustering component: DBSCAN and MeanShift. 
The results of Table~\ref{tab:clustering} show similar performance of these two variations under the proposed framework. 
DBSCAN approach is slightly more sensitive, since it can merge neighboring mushrooms if the existence predictions do not provide well-separated regions. 
To this end, we considered the latter approach (MeanShift) as the default option in our work.

\begin{table}[h]
  \centering
  \resizebox{.8\linewidth}{!}{
  \begin{tabular}{@{}lcc@{}}
    \toprule
    clustering & MAP @ 25\% IoU & MAP @ 50\% IoU\\
    \midrule
    DBSCAN & 99.20\% & 81.48\% \\
    MeanShift & 99.57\% & 81.95\% \\
    \bottomrule
  \end{tabular}
  }
  \caption{Comparison of the two considered instance segmentation approaches, based on different clustering algorithms.}
  \label{tab:clustering}
\end{table}

To have a finer evaluation of the quality of the detections, we also calculated scale and orientation errors for every prediction above 25\% IoU threshold. 
The results are summarized in Table~\ref{tab:errors}, where we also report the corresponding errors for the best performing RANSAC variants. 
We also applied an ICP fine-tuning algorithm, as a template registration approach, over the proposed method.
Notably, even though the proposed method is an one-shot approach, it achieves very low orientation error, while an ICP fine-tuning step decreases performance.

\begin{table}[h]
  \centering
  \resizebox{\linewidth}{!}{
  \begin{tabular}{@{}lccc@{}}
    \toprule
    metric & RANSAC-based & Proposed & Proposed+ICP  \\
    \midrule
    scale relative error & 6.7\% & \textbf{4.8\%} & 12.42\% \\
    \midrule
    cosine similarity & 0.9643 & \textbf{0.9973} & 0.9828\\
    theta error & 15.35$^\circ$ & \textbf{4.22$^\circ$} & 10.63$^\circ$\\
    \bottomrule
  \end{tabular}
  }
  \caption{Scale and angle errors for the best-performing RANSAC variant, the proposed pipeline and the proposed pipeline along with an ICP fine-tuning registration step. The reported mean errors are calculated over detections above 25\% IoU threshold.}
  \label{tab:errors}
\end{table}

\subsection{Qualitative Evaluation over Real Data}

\begin{figure}[h]
    \centering
    \begin{tabular}{cc}
     \includegraphics[width=.34\linewidth]{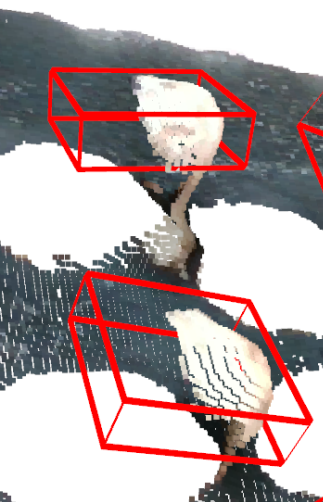} &
     \includegraphics[width=.34\linewidth]{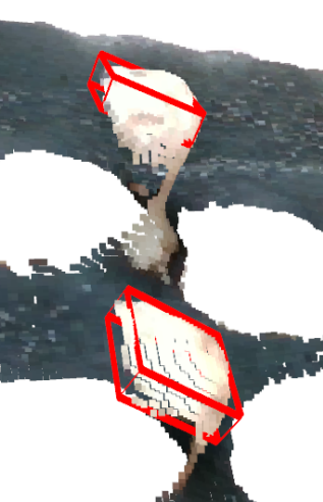} 
    \end{tabular}
    \caption{Examples of pose estimation for a single-view point cloud of real data. We compare two approaches: proposed (left) vs proposed \& ICP fine-tuning (right). Note how ICP decreases the quality of the results.} 
    \label{fig:single-view}
\end{figure}

Finally, in Figure~\ref{fig:vis_bboxes} we provide examples of the effectiveness of our method in real data, hinting a good adaptation despite the synthetic-to-real domain gap.
Notable examples of this adaptations can be found in bottom-middle image, where 3 mushrooms are very close together, as well as the in the top-right image, which corresponds to a single-view point cloud.
Regarding the latter, the proposed synthetic generation simulates single-view scenarios due to the hidden point removal step.
To this end, the proposed method can adapt to cases where only a part of the mushroom is present and extrapolate its pose, contrary to the  aforementioned RANSAC and ICP alternatives.
This exact behavior can be clearly seen in the zoomed-in depiction of Figure~\ref{fig:single-view}, where we compare two different pose estimation approaches (proposed vs proposed \& ICP fine-tuning of Table~\ref{tab:errors}) for a single-view point cloud of real data. 
We can observe that the ICP process tries to ``fit what it sees" that leads to erroneous estimations.

\section{Conclusions}
In this work, we proposed a mushroom segmentation and pose estimation pipeline based on a
lightweight, single-pass network that process 3D point clouds and provides pose-related predictions for each point. These set of predictions is referred to as implicit pose encoding and proved to be critical for training a well-performing system. 
To train such network we face the issue of 3D annotated data shortage, especially for our task. To this end, we also created a synthetic dataset of point cloud mushroom scenes. 
To validate the effectiveness of our method we provided quantitative comparisons over synthetic data, as well as qualitative examples over real multi-view point clouds, obtained from multiple depth sensors.

\section*{Acknowledgments}
We thank our collaborators Teagasc, Ireland for their technical knowledge and expertise in commercial mushroom production and harvesting. This research has received funding from the European Union’s Horizon 2020 research and innovation programme under grant agreement no. 101017054 (project: SoftGrip).

{\small
\bibliographystyle{ieee_fullname}
\bibliography{egbib}
}

\end{document}